# A Natural Language Processing Pipeline of Chinese Free-text Radiology Reports for Liver Cancer Diagnosis

Honglei Liu[1,2], Yan Xu[3], Zhiqiang Zhang[1,2], Ni Wang[1,2], Yanqun Huang[1,2], Yanjun Hu[3], Zhenghan Yang[3], Rui Jiang[4]*, Hui Chen[1,2]*

[1]School of Biomedical Engineering, Capital Medical University, Beijing 100069, China
[2]Beijing Key Laboratory of Fundamental Research on Biomechanics in Clinical Application, Capital Medical University, Beijing 100069, China
[3]Department of Radiology, Beijing Friendship Hospital, Capital Medical University, Beijing, 100050, China
[4]Ministry of Education Key Laboratory of Bioinformatics; Bioinformatics Division, Beijing National Research Center for Information Science and Technology; Department of Automation, Tsinghua University, Beijing 100084, China.

Corresponding author: Rui Jiang (e-mail: ruijiang@tsinghua.edu.cn). Hui Chen (e-mail: chenhui@ccmu.edu.cn).

The work is supported by grants from National Natural Science Foundation of China (No. 81701792 and No. 81971707), and National Key Research and Development Program of China (No. 2018YFC0910404).

**ABSTRACT** Despite the rapid development of natural language processing (NLP) implementation in electronic medical records (EMRs), Chinese EMRs processing remains challenging due to the limited corpus and specific grammatical characteristics, especially for radiology reports. In this study, we designed an NLP pipeline for the direct extraction of clinically relevant features from Chinese radiology reports, which is the first key step in computer-aided radiologic diagnosis. The pipeline was comprised of named entity recognition, synonyms normalization, and relationship extraction to finally derive the radiological features composed of one or more terms. In named entity recognition, we incorporated lexicon into deep learning model bidirectional long short-term memory-conditional random field (BiLSTM-CRF), and the model finally achieved an F1 score of 93.00%. With the extracted radiological features, least absolute shrinkage and selection operator and machine learning methods (support vector machine, random forest, decision tree, and logistic regression) were used to build the classifiers for liver cancer prediction. For liver cancer diagnosis, random forest had the highest predictive performance in liver cancer diagnosis (F1 score 86.97%, precision 87.71%, and recall 86.25%). This work was a comprehensive NLP study focusing on Chinese radiology reports and the application of NLP in cancer risk prediction. The proposed NLP pipeline for the radiological feature extraction could be easily implemented in other kinds of Chinese clinical texts and other disease predictive tasks.

**INDEX TERMS** Natural Language Processing, radiology reports; information extraction, computer-aided diagnosis, BiLSTM-CRF

## I. INTRODUCTION

Massive electronic medical records (EMRs) are potentially valuable clinical sources for research for improving clinical care and support [1, 2]. In the current digital age, machine learning-based algorithms play a powerful role in data mining, which is useful in applications such as clinical decision-making, disease computer-aided diagnosis, and management [3, 4].

As an important EMRs component, the radiology report is a primary method of communication between radiologists who interpret the image and physicians who make the final diagnosis. Radiological diagnosis is frequently formulated by relying on physicians' experience, which may lead to limited accuracy and efficiency [5]. With the rapid growth of clinical big data, applying machine learning methods to process medical texts becomes executable. Extracting clinically relevant information from radiology reports has great importance in terms of advancing radiological research and clinical practice [6], although significant challenges still exist, mainly due to the free form of most reports [7]. Natural language processing (NLP) is a multistep process comprised of statistical and linguistic methods that can mine information from unstructured texts, which are then formed into a standardized structured format (i.e., a fixed collection of text features). NLP-based feature extraction has advantages for massive text processing compared with time-consuming manual extraction flow. Hence, NLP-based feature extraction has been effectively used in radiology for diagnostic surveillance, cohort building, quality assessment, and clinical support services [8-11]. Nevertheless, previous NLP studies of radiology reports primarily focused on documents written in English. With the rapid growth of clinical data in China, information extraction from vast amounts of Chinese radiology reports has become a meaningful task that has both theoretical and practical significance. Due to the limitations of the related corpus, NLP on Chinese clinical texts remains challenging [12, 13].

Compared with structured text, free text is more natural and expressive in the record of the clinical events. To facilitate the application of clinical texts, information-mining research using NLP, which could automatically extract entities, events, and relations, is necessary. During the NLP workflow, such as semantic analysis and syntactic analysis, a lexicon of words with definitions and synonyms is useful. Several tools and systems could provide such support. For example, the Unified Medical Language System (UMLS) Metathesaurus [14] includes synonymous terms and specific semantic roles for each concept and relationships between concepts. Other useful lexicons and ontologies include RadLex® [15], which is a specialized radiological lexicon including imaging techniques.

Named entity recognition (NER) is a fundamental NLP task, which could be seen as a sequence labeling tasks. Clinical NER is a critical task for information extraction from EMRs. Clinical NER aims to identify and classify terms in EMRs, such as diseases, symptoms and examination types [16]. In last decade, a number of methods proposed for clinical NER, which could mainly be divided into two categories: knowledge-driven methods based on rules and corpus, and data-driven machine learning methods. Machine learning methods include Hidden Markov Models (HMM), Maximum Entropy Markov Models (MEMM), Conditional Random Field (CRF) and so on. Recently, deep learning models were introduced into NER to improve the performance. Of all the deep learning models, bidirectional long short-term memory (BiLSTM) is a variant of the Recurrent Neural Network (RNN), which could effectively capture long-range related information effectively in NER task. By splitting the neurons into two directions of a text sequence, BiLSTM could learn forward and backward information of input words, Furthermore, BiLSTM with CRF (BiLSTM-CRF), proved its validity that outperformed the traditional models especially in Chinese clinical NER tasks [17-21].

After information extraction to obtain the structured features, NLP can be further implemented on clinical tasks, such as disease studies [22-24], drug-related studies [25, 26], and clinical workflow optimization [27]. Computer-aided diagnosis is an important research field in disease study, which aims to use computer algorithms to provide physicians a reference for disease diagnosis. Studies have investigated many diseases to date, such as hepatocellular cancer [22], colorectal cancer [28], pancreatic cancer [24], and celiac disease [29]. Wu et al. developed Med3R using a deep learning model that successfully provided a comprehensive aided clinical diagnosis service on EMRs [30]. Liang et al. applied an automatic NLP system to provide clinical decision support and achieved a high diagnostic accuracy in pediatric diseases. The NLP system could extract key concepts and then transformed them into reformatted data in query-answer pairs [31].

Due to the limitation of Chinese EMRs corpus, NLP systems in clinical information extraction and application are challenging, which probably leads to a poor performance based on the general corpus. Therefore, corpus annotating and lexicon building are necessary for NLP in specific clinical applications. Recently, there are increasing numbers of studies on broader NLP element tasks in Chinese EMRs, such as NER [32] and speculation detection [33].

For radiology reports, NLP has been utilized for identifying biomedical concepts [34], extracting recommendations [35], determining the change level of clinical findings [36], and so on. Machine learning methods are widely used today for other clinical applications. For example, Bahl et al. developed a random forest method to predict high-risk breast lesions using textual features [37]. Using IBM Watson's NLP algorithm, Trivedi et al. developed a classifier to automatically assign the intravenous contrast use based on magnetic resonance imaging reports [38].

Although there are some studies based on Chinese clinical texts in NLP fundamental tasks, higher-level tasks and applications are limited, especially for research on radiology reports. Building a comprehensive NLP pipeline for information extraction from Chinese radiology reports has great importance for further NLP research. In this study, we designed an NLP pipeline that could extract clinically relevant radiological features from abdominal computed tomography (CT) radiology reports written in Chinese. Unlike other Chinese medical text information extraction systems, our study extracted all possible radiological features which were consisted of entities based on specific rules. The number and content of radiological features were not determined in advance in our study. Then we used Lasso (least absolute shrinkage and selection operator) for radiological feature selection. Lasso is a regression analysis method for the feature selection to improve the prediction performance and interpretability. Imposing L1 penalty on the feature vector, the Lasso method encouraged to use only a subset of the overall features rather than all of them [39].

The NLP pipeline was comprised of NER, synonyms normalization, and relationship extraction. In consideration of the language characteristics of Chinese, we manually collected a lexicon, containing words and synonym lists. We incorporated the lexicon into BiLSTM-CRF to improve the performance of NER tasks. Typically, patients with liver cancer are likely to be diagnosed with symptoms of advanced disease. Moreover, the diagnosis of liver cancer via early examination, such as radiological examination, is necessary [40]. Therefore, in terms of implementation, we applied different machine learning algorithms to liver cancer prediction using the radiological features extracted by the NLP pipeline.

## II. MATERIALS & METHODS

*Dataset*

Abdominal CT radiology reports were collected from a tertiary hospital in Beijing, China, between 2012 and 2018. The study and data use were approved by the Human Research Ethics Committees of Beijing Friendship Hospital, Capital Medical University, Beijing, China. All identifying information was removed to protect patient privacy. All radiology reports were unstructured and written in Chinese. According to the content, the radiology report included the Type of examination, Clinical history, Comparison, Technique, Findings, and Impressions. In the findings section, a radiologist listed the observations regarding each area of the body examined. Whether and how the area was normal, abnormal or potentially abnormal was recorded. The impressions section contained a diagnosis indicated by a radiologist when combining the radiological findings and clinical history. The NLP pipeline in this study was applied to the section of radiological findings.

Of all the patients, 480 were diagnosed with liver cancer based on both the section of impressions and annotations by experienced radiologists. We further randomly selected 609 reports of 609 patients with the diagnosis of liver cirrhosis, liver cysts, hepatic or hemangioma (Supplementary Figure 1).

*The NLP pipeline*

Figure 1 shows the overview of the computer-aided diagnosis framework that consisted of lexicon building, NLP and disease classifier. NLP was performed to extract radiological features with terms from the radiology reports. Features in training reports were reduced to a smaller subset by Lasso, and then were input into machine learning models.

### 1) LEXICON BUILDING

The whole framework for feature selection and extraction was initialized with lexicon construction. In reports with and without a liver cancer diagnosis, a small number of reports (approximately 3% of overall data) were sampled randomly for generating the lexicon by manual reading. Another subset of radiology reports (approximately 1% of the overall data in the rest of the data) was sampled randomly to further manually integrate the lexicon. We then invited an experienced radiologist to proofread the lexicon. We randomly selected five reports from the rest to validate the completeness of the lexicon.

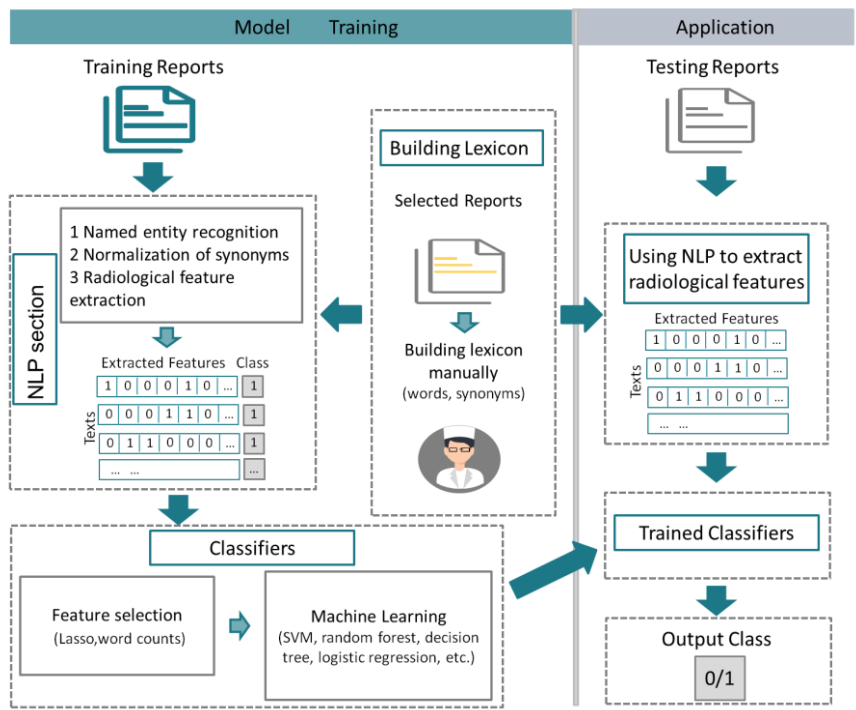

**FIGURE 1.** Overview of the natural language processing pipeline

TABLE 1. AN EXAMPLE OF LEXICON FEATURES AND TAGS

| Character Sequence | 肝 | 脏 | 形 | 态 | 大 | 小 | 正 | 常 | , | 轮 | 廓 | 规 | 整 |
|---|---|---|---|---|---|---|---|---|---|---|---|---|---|
| Lexicon Feature Sequence | B | E | None | None | None | None | None | None | None | B | I | I | E |
| Entity Annotation Tag Sequence | B-L | E-L | B-M | I-M | I-M | I-M | I-M | E-M | O | B-M | I-M | I-M | E-M |
| Entity Type | Location | | | Morphology | | | | | | | Morphology | | |

\* The B, I, E, O tags indicated the Begin, Inside, End, and Outside of one word. The B-L, I-L tags indicated the beginning and inside of the entity type [Location], respectively. The B-M, I-M tags indicated the beginning and inside of the entity type [Morphology], respectively

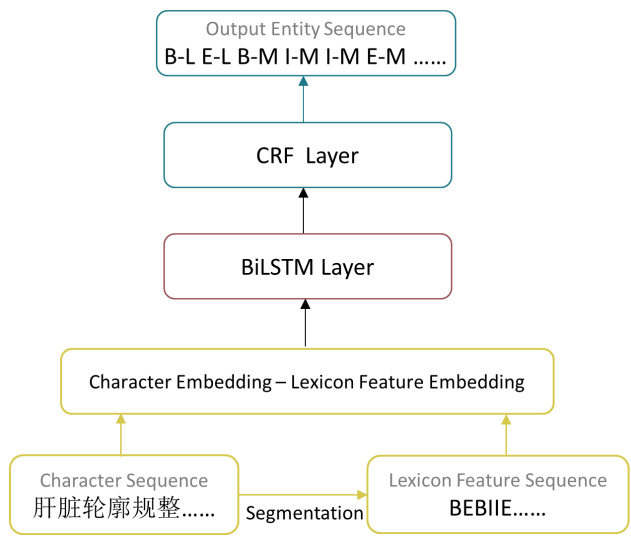

**Figure 2.** The architecture of BiLSTM-CRF model. The B-L/B-M, I-L/I-M tags indicates the beginning and inside of the entity type [Location]/[Morphology], respectively. The B, I, E tags indicates the Begin, Inside, and End of one word.

After segmentation using the forward maximum matching algorithm, we found that the current lexicon could cover all of the clinically relevant words in these reports. The specialized lexicon containing clinical terms and lists of synonyms were built based on prior clinical knowledge and Chinese grammatical characteristics. Synonyms involved different locations of the liver and different presentations of items, such as "low density", "irregular".

2) RADIOLOGICAL FEATURE EXTRACTION
In this study, we performed a deep learning algorithm, i.e., the BiLSTM-CRF model for clinical NER task. The entity types in radiology reports included [Location] (e.g., 肝脏 (liver)), [Morphology] (e.g., 轮廓规整 (regular contour)), [Density] (e.g., 密度不均匀 (nonhomogeneous)), [Enhancement] (e.g., 动脉期 (arterial phase)), and [Modifier] (e.g., 结节状 (nodular)). The goal was to assign the BIEOS (Begin, Inside, End, Outside, Single) tags to each Chinese character according to different entity types. The deep learning model contained three layers, the word embedding layer, BiLSTM layer, and CRF layer (Figure 2).

The goal of word embedding was transforming the discrete Chinese characters into a vector representation from a large amount of unannotated text. Despite the entity annotation tags, we also added lexicon features into word embedding. The lexicon was used for word segmentation by the classic forward maximum matching algorithm. We then generated the lexicon feature sequence according to the segmentation result. BIEOS tags were then annotated according to segmentation results, resulting in the lexicon feature sequence (Table 1). We got the word embedding results of character sequence and lexicon feature sequence, respectively, and then integrated the embedding results together to represent the character sequence (Figure 2). The Word2Vec was used for word embedding after being pre-trained on the Chinese Wikipedia data. The BiLSTM layer could capture the dependencies of adjacent tags and learn forward and backward information of input Chinese characters.

The BiLSTM model was trained by Adam (adaptive moment estimation) optimization algorithm, which was widely used in deep learning. We set the number of hidden units to 100, and the optimizer to Adam. Constraints existed in the sequence labeling step since adjacent tags had dependencies. For example, the entity should start with B tag, and I tag must follow the B tag. After the BiLSTM step, we applied the CRF model to compute the optimal sequence tags.

In the word-level normalization of synonyms, entities with the same meaning were unified into a single word according to the synonym lists generated previously. We then extracted symptom information among the single entities towards the computer-aided diagnosis. Each report was divided into a series of sentences by a full stop. The sentences were further divided into several parts if more than one entity [Location] occurred. As described in Table 2, several rule-based patterns were then designed to extract relations according to semantic comprehension, syntactic structure, and knowledge-based characteristics. All patterns started with an entity [Location], followed by the combination of other entities in each sentence or part. Entity [Morphology] only represented the morphology description for different locations. Entity [Modifier] referred to the modifier of entity [Enhancement] or [Density]. If the entities [Enhancement] and [Density] occurred at the same time, the feature

([Location]+[Enhancement]+[Density]+Others (either [Modifier] or [Morphology])) would be recognized into several features all starting with the same [Location] ([Location]+[Enhancement]+Others or [Location]+[Density]+Others or [Location] +Others). We scanned all the satisfactory patterns according to the above rules. A feature extraction example is shown in Figure 3. These lists of radiological features were subsequently used to build prediction models for liver cancer.

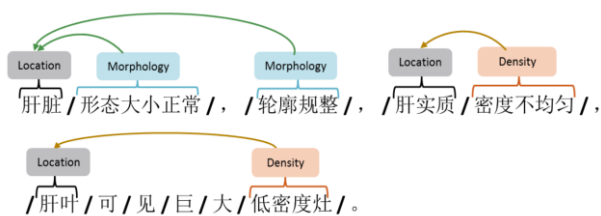

**A sentence from the imaging findings of a radiology report:**
肝脏形态大小正常，轮廓规整，肝实质密度不均匀，肝右叶可见巨大低密度灶。
(The liver is normal in size and shape, and the contour is regular. The liver parenchyma is nonhomogeneous. Right lobe of liver has low density area.)

**A. Named entity recognition, normalization of synonyms and relation extraction**

**B. Radiological features**

| 1 肝脏 / 形态大小正常 | (liver / normal in size and shape) |
| 2 肝脏 / 轮廓规整 | (liver / contour is regular) |
| 3 肝实质 / 密度不均匀 | (liver parenchyma / nonhomogeneous) |
| 4 肝叶 / 低密度灶 | (liver lobe / low density area) |

**FIGURE 3.** An example of feature extraction in a sentence from a radiology report. Texts in parentheses are the corresponding English translations.

TABLE 2. PATTERNS SUMMARIZED TO EXTRACT FEATURES ACCORDING TO THE WORD'S ENTITY TYPE

| Entity pattern | Example |
| --- | --- |
| **Location + Density** | 肝脏+低密度影 (liver + low density) |
| **Location + Enhancement** | 肝脏+增强扫描未见强化 (liver + enhancement scan showed no enhancement) |
| **Location + Enhancement + Modifier** | 肝门+动脉期+结节状强化 （porta hepatis + arterial phase + nodular enhancement） |
| **Location + Density + Modifier** | 肝脏+低密度灶+边界清晰 (liver + low density area + clear boundary) |
| **Location + Morphology** | 肝脏+形态大小正常 (liver + normal in size and shape) |

### 3) PREDICTIVE MODELS
Itemized features derived from the previous steps were binary, representing the absence or presence of a certain feature (Figure 1). They served as the input of the classifier for liver cancer prediction. The classifier output was also binary, indicating whether the patient was diagnosed with liver cancer or not.

We introduced Lasso for the feature selection. Features selected by Lasso were further used by the test reports. We used the binomial distribution for Lasso logistic regression due to the binary response (whether diagnosed with liver cancer or not) in this study.

Machine learning-based classifiers, including the decision tree, random forest, support vector machine (SVM) and logistic regression were then built for the liver cancer prediction, respectively. With good interpretability, logistic regression is usually used to explain the relationship between the independent variables and the binary dependent variable. A decision tree can be considered as a set of if-then rules, which describes the process of instances classification based on trees. The prediction model and results generated by a decision tree are easy to understand [41]. Random forest is an ensemble learning method constructed with a multitude of decision trees, and random forest usually gets higher performance than a single decision tree [42]. Based on the structural risk minimization principle, SVM is a robust model for prediction problems by maximizing the margin. Different types of kernels can be chosen to solve both linear and non-linear problems [43], while a linear kernel was used in this study.

Fivefold cross-validation was employed when assessing and comparing the predictive models. Performance measures used in this classification study included recall (also called sensitivity), precision and F1 score. To rank the radiological features associated with the diagnosis of liver cancer, feature importance score was computed by Gini impurity in the random forest method. Gini impurity is a measurement of the probability that a sample is classified incorrectly in tree-based models without a specific feature.

### III. RESULTS
We finally collected 831 words and 48 lists of synonyms in the lexicon. In NER task, we recognized the entity types [Location], [Morphology], [Density], [Enhancement], and [Modifier] using BiLSTM-CRF model, and compared the recognition results of the models with or without lexicon (Table 3). For BiLSTM-CRF model with the lexicon, word embedding results of character sequence and lexicon feature sequence were integrated to represent the character sequence. While for the model without lexicon, the lexicon feature sequence was not included. Character sequence was represented by its word embedding result only. For all the entity types, our proposed model with lexicon achieved the performance with a precision of 92.35%, a recall of 93.66%, and an F1 score of 93.00%. In addition, the lexicon features bring an improvement of 1.74% in precision, 2.72% in recall and 2.22% in F1 score for the basic BiLSTM-CRF model without lexicon. Except entity [Density], model with lexicon got higher performance than the model without lexicon.

A radiological feature was not considered universal if its frequency was too low. We invited a radiological expert to review the features with frequency less than 0.5% (i.e., occurred less than five times in all the 1080 reports). We didn't find clinically meaningful features with a frequency less than three, such as "肝脏/结构" (liver/structure) and "肝脏/填充" (liver/filling). Among features with a count equal to three, there existed clinically meaningful features, such as "肝脏/结构紊乱" (liver/disorder structure) and "肝脏/边缘光整" (liver/finishing edge). Therefore, we set the frequency to 0.3%, that is, remaining the features occurred more than twice. We finally got 109 features to formulate the feature vectors (Supplementary Table 1). The features described the normality and abnormality of liver morphology, liver density, and enhancement. They also contained morphology of other locations, such as abdominal pelvic and portal vein. According to the presence or absence of each feature, every radiology report was represented by a 0-1 vector in the feature vector space. The statistics of the extracted radiological features are shown in Table 4. There were six features with a proportion higher than 30% of all the reports. The top two features with high proportion were associated with liver morphology, which were usually required to be recorded in every radiology report in the routine radiology practice.

TABLE 3. NAMED ENTITY RECOGNITION RESULTS USING BiLSTM-CRF.

| BiLSTM-CRF | Entity Type | Precision (%) | Recall (%) | F1 Score(%) |
|---|---|---|---|---|
| **Without Lexicon** | [Location] | 93.55 | 93.55 | 93.55 |
| | [Morphology] | 93.17 | 94.31 | 93.74 |
| | [Density] | 80.39 | 89.91 | 84.89 |
| | [Enhancement] | 88.44 | 84.14 | 86.24 |
| | [Modifier] | 80.40 | 78.83 | 79.61 |
| | All | 90.61 | 90.94 | 90.78 |
| **With Lexicon** | [Location] | 94.41 | 96.48 | 95.44 |
| | [Morphology] | 96.07 | 96.00 | 96.04 |
| | [Density] | 80.21 | 90.06 | 84.85 |
| | [Enhancement] | 88.95 | 88.49 | 88.72 |
| | [Modifier] | 84.08 | 78.33 | 81.10 |
| | All | 92.35 | 93.66 | 93.00 |

All the models got a relatively high performance (Table 5), and Lasso worked efficiently in performance improvement. All four classifiers with Lasso-based feature reduction got a higher F1 score compared with classifiers without such feature dimension reduction. The highest F1 score of 86.97% was seen in the random forest model, whose recall was also the highest (86.25%). Compared with random forest, the logistic regression got a lower F1 score but a higher precision. All the evaluation indicators of random forest were higher than those of a decision tree. After Lasso being applied, the performance of SVM and logistic regression improved greatly. F1 score increased by 7.13% for logistic regression and 4.01% for SVM. However, the decision tree and random forest were not sensitive to the reduced input features, with F1 score improvements of only 2.35% and 2.02%, respectively (Table 5). The feature importance scores of all radiological features were derived from random forest and the top ten features associated with the liver cancer diagnosis were shown in Supplementary Figure 2. The top three features included the existence of clear enhancement and low density, and the regular state of liver shape.

TABLE 4. RADIOLOGICAL FEATURES WITH A COUNT GREATER THAN 300.

| Radiological Features | Count | Proportion in all the reports | Proportion in all the features |
|---|---|---|---|
| 肝脏 / 形态大小正常 (liver / normal in size and shape) | 551 | 50.60% | 8.96% |
| 肝脏 / 轮廓规整 (liver / contour is regular) | 483 | 44.35% | 7.85% |
| 肝裂 / 无增宽 (hepatic fissures / no broadening) | 385 | 35.35% | 6.26% |
| 肝脏 / 低密度影 (liver / low density) | 373 | 34.25% | 6.06% |
| 肝叶 / 比例如常 (liver lobe / normal proportion) | 370 | 33.98% | 6.01% |
| 肝门 / 未见异常 (porta hepatis / regular) | 329 | 30.21% | 5.34% |

TABLE 5. PERFORMANCE OF DIFFERENT MACHINE LEARNING MODELS FOR LIVER CANCER DIAGNOSIS

| Predictive model | Precision (%) | Recall (%) | F1 Score (%) |
|---|---|---|---|
| **Without Lasso** | | | |
| Logistic Regression | 80.74 | 77.71 | 79.10 |
| Decision Tree | 80.39 | 76.88 | 78.59 |
| Support Vector Machine | 70.59 | 80.00 | 75.00 |
| Random Forest | 87.78 | 82.29 | 84.95 |
| **With Lasso** | | | |
| Logistic Regression | 87.72 | 84.79 | 86.23 |
| Decision Tree | 83.26 | 78.75 | 80.94 |
| Support Vector Machine | 81.23 | 76.88 | 79.01 |
| Random Forest | 87.71 | 86.25 | 86.97 |

## IV. DISCUSSION

Liver cancer is a substantial economic burden for both patients and the government in China. Limited by the diagnostic technology, many patients are diagnosed at the stage of terminal liver cancer, resulting in a much poorer prognosis in China compared with that of developed countries [44]. Therefore, the early diagnosis of liver cancer by the benefit of informative examination, such as radiological examination, has great significance [40, 45]. Free text-based texts could not be directly used in machine learning algorithms. Therefore, NLP methods for the extraction of structured features were essential. For clinical texts, NLP was implemented in disease study areas, especially for the category of neoplasms [1].

There was limited corpus for Chinese EMR processing, especially for the research of Chinese radiology reports. Some research groups have built small-scale, disease-specific corpus for their research [31, 46]. Therefore, building the corpus manually for our research was necessary. Our work can be used as a reference in similar applications of Chinese EMRs and will contribute to a possible future large-scale Chinese EMR corpus. In this work, we constructed a lexicon from a small proportion of radiology reports randomly sampled in the overall dataset. This radiological lexicon, rather than a general dictionary, was then used in the subsequent study. The lexicon was manually collected and annotated by some experienced radiologists based on their prior clinical knowledge and Chinese grammatical rules. Since the lexicon was built manually, the more reports reviewed, the more manual labor used. After balancing the labor and the quality of the lexicon, we finally chose to use 4% of all the reports. Different from English, Chinese has its own specific semantic characteristics and grammatical rules, especially in the medical domain. Chinese text has more flexibility in word combinations. For example, the word 肝脏 (English: liver), belonging to entity [Location], could also be written as a specific segment of the liver in radiology reports, such as "肝S8", "肝S3", or just one character, "肝". Therefore, the constructed lexicon included a list of synonyms to unify different presentations and different sections of 肝脏 into a single word. The synonyms also contained other Chinese expressions, such as negative words. The lexicon only took clinically relevant words into consideration. As a result, other words remained unique characters and would be ignored during information extraction. With the manually build lexicon, the performance of NER and normalization of synonyms were greatly improved. Furthermore, we built the lexicon from only a small proportion of radiology reports. Therefore, this pipeline could be used as a reference in similar applications of Chinese EMRs.

In the consideration of the characteristics of radiology reports, we annotated five entity types and designed deep learning-based BiLSTM-CRF model for the NER task. The BiLSTM-CRF model has outperformed the traditional models and achieved the state-of-art results in Chinese NER tasks. Some studied introduced dictionaries into deep neural networks and got higher performance than the reference model [18, 47]. Chinese and English clinical texts have different characteristics in linguistic traits and writing styles. In Chinese, a token is a character, while in English a word is usually a token. Therefore, in NER task, model with segmentation result could improve the performance. We introduced Chinese lexicon features into the word embedding step based on the manually collected lexicon. After word representation with the lexicon information, clinical knowledge could be added into the deep learning model and provide valuable information when dealing with rare cases. Furthermore, an entity could be seen as one word or several words, therefore the segmentation results by lexicon could introduce boundary information of entities. Compared with the model without lexicon, BiLSTM-CRF with lexicon could get higher performance in all the types except [Density]. The F1 score of entity type [Location], [Morphology], [Enhancement] and [Modifier] increased by 1.89%, 2.30%, 2.48%, and 1.49%, respectively.

We designed five patterns (i.e., entity combinations) for the radiological feature extraction. The extracted itemized features could present the meaning of corresponding sentences. Although the listed patterns and entity annotation could restrict the number of word combinations, the features still had a high dimension. Screened by word count and the Lasso method, the extracted features decreased to a tiny amount, which was a relatively limited number compared with the free-text. As presented in Table 4, the normal morphology of different locations had the highest counts. The main reason for this may be that the morphology of some locations, such as liver, liver lobe, and porta hepatis, should be recorded in every radiology report no matter whether the patient had liver disease or not.

Among the overall 1089 reports, almost half features (50/109) appeared in less than 10 reports, resulting in a rather sparse feature matrix. Therefore, we chose Lasso for feature selection. With the derived radiological features containing terms, all four machine learning models had good performance with an F1 score higher than 75% with Lasso (Table 5), where random forest achieved the highest F1 score. All the evaluation indicators of random forest were higher than the decision tree since the random forest was an ensemble method constructed by a large number of decision trees. Random forest could realize the feature selection. Therefore, it was not sensitive to the reduced input features. The precision, recall and F1 scores of models with and without Lasso were close to each other. In Lasso, variable selection and complexity adjustment were carried out while fitting the generalized linear model. Logistic regression was a kind of the generalized linear model and was used in Lasso. Therefore, the performance of logistic regression with Lasso improved greatly. For the two classification models with an overall high performance, logistic regression achieved a higher precision but lower recall than the random forest, meaning that the logistic regression-based classification model had a higher positive predictive value and was more likely to provide a false negative prediction. We could see that different performance occurred with the same features using different classifiers. In clinical application, lower recall represented a higher under-diagnosed rate, which was not beneficial for disease screening. In contrast, lower precision showed lower prediction reliability, leading to a lower clinical application value. Therefore, the four machine learning methods adapted to various application requests. Logistic regression had the highest reliability and random forest had the highest completeness in liver cancer prediction. We could conclude that the structured features extracted by the NLP pipeline have obtained effective information from the original reports in this study.

Through the analysis of misjudgment samples, we identified some patients who were diagnosed with liver cirrhosis were easily classified as liver cancer since some radiological features of liver cirrhosis were close to liver cancer. Patients with liver cirrhosis had the potential to progress to cancer [40, 45]. Therefore, our results could be an early warning for these patients. Another reason for the incorrect classification may be the radiological features omitted during NLP extraction. Due to the size of the dataset, the missing of clinical terms during lexicon construction was inevitable. Especially due to Chinese grammatical characteristics, the long term could retain the same meaning after the emendation of several characters in unstructured form. Thus, extending the lexicon to cover as much term as possible in free-text radiology reports was a challenging task. We could collect more data for information extraction or more samples for lexicon construction to decrease this kind of error in future studies.

The prediction of cancer and other diseases is an important and significant application of medical language processing. Extracted features from EMRs could be part or all of the features for classifier input. Studies of cancer prediction using administrative data and EMRs have been published [22]. In recent years, there were also several studies of disease evaluation using NLP on Chinese clinical data [30]. In contrast with these studies of NLP applications, the NLP pipeline in this work focused on features extracted only from texts, which could represent the whole free text in further applications. Compared with free text-based systems, the structured features extracted in this study had good interpretability, and could be seen as the diagnosis evidence. Furthermore, compared with previous works which extracted isolated entities, our work extracted the radiological features that consisted of several entities, which had more implications in radiology.

To get the radiological features strongly associated with the liver cancer diagnosis, we ranked the features by feature importance score computed by Gini impurity derived from the random forest method. The clear enhancement state of liver had the highest feature importance score which coincided with clinical knowledge [40]. Several top features were important and basic risk factors in liver disease diagnosis.

The NLP pipeline could extract radiological features automatically, which were inputted into the diagnosis model. The diagnosis model could provide diagnosis advice to clinicians. This model was currently a prototype. In future research, with more data and annotation resources, we hope to refine the model for application in clinical practice. Diagnostic results derived from our model are expected trustable and acceptable since it had proved to get high performance. The diagnostic decisions derived from our model were transparent with the extracted radiological features, and the top ten features associated with the liver cancer diagnosis were consistent with existing clinical knowledge for liver cancer diagnosis. Furthermore, we had conducted a previous investigation on how a neural network-based computer-aided diagnosis scheme could help radiologists make diagnostic decisions [48]. It showed that radiologists, especially junior radiologists with limited practical experience, were more likely to trust the computer-aided diagnosis scheme, and their diagnostic ability improved a lot.

Although our pipeline has been shown to have high performance in liver cancer diagnosis, limitation still exists. The lexicon construction was based on limited annotation resources from one hospital. Hence, some clinical key terms had a risk of omission, and the performance of some NLP procedures might weaken across different NLP tasks.

## V. CONCLUSIONS

This study described an NLP pipeline of Chinese free-text radiology reports for liver cancer diagnosis. We incorporated lexicon into deep learning model BiLSTM-CRF to improve the NER performance. Our model achieved a high performance both in the NER and liver cancer prediction. This work was a comprehensive study of a liver cancer computer-aided diagnosis model using the NLP method based on Chinese radiology reports. The proposed NLP pipeline could be generalized to the lexicon construction of other diseases and other kinds of clinical texts in Chinese. Furthermore, the radiological feature extraction method will expect to be an important step towards the use of massive Chinese clinical data for health research.

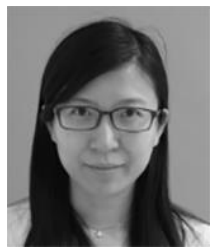

**HONGLEI LIU** received the B.S. degree in electronic information engineering from Central South University, in 2010, and the Ph.D. degree in control science and technology from Tsinghua University, in 2016. From 2013 to 2014, she was a Visiting Scholar with Center for Systems and Synthetic Biology, University of California, San Francisco (UCSF). Since 2016, she has been a Lecture with School of Biomedical Engineering, Capital Medical University. Her research interests include medical information, medical natural language processing.

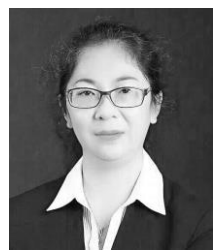

**YAN XU** received the M.D. degree in imaging medicine and nuclear medicine from Capital Medical University, in 2013. She is now a Chief Physician with Beijing Friendship Hospital, Capital Medical University. Her research interests include imaging diagnosis of chest diseases.

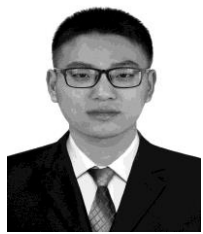

**ZHIQIANG ZHANG** received the B.S. degree in electronic information science and technology from Southwest Jiaotong University, in 2018. He is currently pursuing a degree in Biomedical Engineering with Capital Medical University. His research interests include natural language processing in electronic health records.

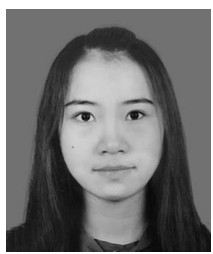

**NI WANG** received a Bachelor's degree in Biomedical Engineering from Capital Medical University in 2017. She is currently pursuing the Ph.D. degree in Biomedical Engineering with Capital Medical University. Her research interests include data mining and secondary use of electrical medical records data.

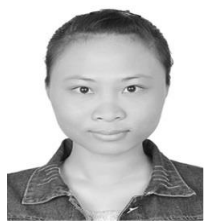

**YANQUN HUANG** received the B.S. degree in Biomedical Engineering from Capital Medical University, in 2018. She is currently pursuing a MS degree in Biomedical Engineering with Capital Medical University. Her research interests include representation learning in electronic health records.

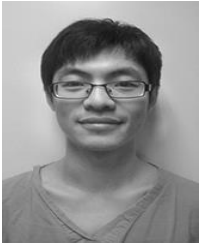

**YANJUN HU** received the B.S. degree in Biomedical Engineering from Capital Medical University, in 2009. He is currently a Junior Engineer with Beijing Friendship Hospital, Capital Medical University. His research interests include data mining of Electrical Medical Records data.

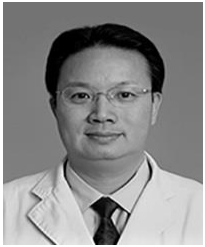

**ZHENGHAN YANG** received the M.D. degree in imaging medicine and nuclear medicine from Peking University Health Science Center, in 1999. He is now a Chief Physician with Beijing Friendship Hospital, Capital Medical University. His research interests include imaging diagnosis of abdominal diseases.

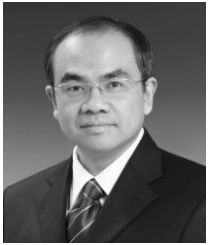

**RUI JIANG** received the B.S. and Ph.D. degree in Automation from Tsinghua University, China, in 1997 and 2002. Since 2007, he joined Tsinghua University and is now an Associate Professor in Department of Automation. His research interests include artificial intelligence and big data of health care.

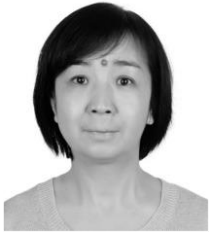

**HUI CHEN** received the Ph.D. degree in Biomedical Engineering from Capital Medical University, in 2009, where she is currently a Professor. Her research interests include secondary use of electrical medical records data, medical informatics and clinical natural language processing.